\documentclass{article}

\usepackage[final]{neurips_2021}

\usepackage[utf8]{inputenc} 
\usepackage[T1]{fontenc}    
\usepackage{hyperref}       
\usepackage{url}            
\usepackage{booktabs}       
\usepackage{amsfonts}       
\usepackage{nicefrac}       
\usepackage{microtype}      
\usepackage{xcolor}         
\usepackage{graphicx}
\usepackage{caption}
\usepackage{subcaption}
\usepackage{dsfont}
\usepackage{amsmath}
\DeclareMathOperator*{\argmax}{argmax} 
\usepackage{floatrow}
\usepackage{float}
\floatstyle{plaintop}
\restylefloat{table}

\title{Have I done enough planning or should I plan more?}

\author{
  Ruiqi He\textsuperscript{1}, Yash Raj Jain, Falk Lieder \\
  Max Planck Institute for Intelligent Systems,  T\"ubingen\\
  \textsuperscript{1} \texttt{ruiqi.he@tuebingen.mpg.de} \\
}

\begin{document}

\maketitle

\begin{abstract}
People's decisions about how to allocate their limited computational resources are essential to human intelligence. An important component of this metacognitive ability is deciding whether to continue thinking about what to do and move on to the next decision. Here, we show that people acquire this ability through learning and reverse-engineer the underlying learning mechanisms. Using a process-tracing paradigm that externalises human planning, we find that people quickly adapt how much planning they perform to the cost and benefit of planning. To discover the underlying metacognitive learning mechanisms we augmented a set of reinforcement learning models with metacognitive features and performed Bayesian model selection. Our results suggest that the metacognitive ability to adjust the amount of planning might be learned through a policy-gradient mechanism that is guided by metacognitive pseudo-rewards that communicate the value of planning.   
\end{abstract}

\section{Introduction}
Humans are frequently confronted with complex problems that require planning and executing long sequences of appropriate actions to reach distant goals. A search tree can be used to depict the space of future actions and consequences. However, such trees grow exponentially as the length of the sequences increases. While current trends in artificial intelligence depend on the exponential increase in computational power, the human mind's cognitive resources are much more limited. Therefore, how is it possible that people are nevertheless able to outperform computers on a wide range of difficult real-world tasks? One critical capacity that enables people to do more with less computation is meta-reasoning, that is reasoning about reasoning \cite{Griffiths2019}. In the context of planning, this means making intelligent decisions about when and how to plan and thereby whether and how to allocate computational resources. In AI research, optimal metareasoning is often regarded to be intractable \cite{Russell1991}. This raises the question of how people are able to solve the apparently intractable metareasoning problem despite their limited computational resources. One intriguing possibility is that people learn an approximate solution 
through trial and error. This idea is known as \textit{metacognitive reinforcement learning} \cite{LiederGriffiths2017,krueger2017enhancing, LiederShenhav2018}.

Previous work found that metacognitive reinforcement learning adapts \textit{which information} people prioritise in their decisions \cite{jain2019people,HeJainLieder2021}. By contrast, in this work, we first aim to show that people are able to learn to adjust \textit{how much} planning they perform to the costs and benefits of planning through a novel experiment. Then to investigate the underlying metacognitive learning mechanism, we fit the resulting data from the experiment with models of metacognitive reinforcement learning and their extensions. We then use Bayesian model selection to test whether human metacognitive learning relies on value-based or policy-gradient mechanisms, whether it is guided by an internally generated metacognitive reward signal, and whether there is a separate meta-control mechanism for deciding when to stop planning. Based on the model selection results, we perform model-based analyses of what differentiates people who could successfully adapt how much planning they perform from participants who struggled to adapt. 

Our new results strengthen the scientific foundation for recreating in machines an essential metacognitive feature of human intelligence, namely the capacity to efficiently allocate limited computational resources \cite{Griffiths2019}. In addition, our results advance the understanding of how people come to make rational use of their limited cognitive resources \cite{lieder2020resource}. Concretely, our findings suggest that this adaptation is achieved through gradual metacognitive learning.



\section{Experiment} \label{sec:Experiment}
To demonstrate that people gradually learn to adapt their amount of planning to its cost and benefits, we designed an experiment using the Mouselab Markov Decision Process (MDP) paradigm \cite{Callaway2017}. 

\subsection{The Mouselab MDP paradigm}
To externalise people's amount of planning, we leverage an experimental paradigm that makes people's behaviour highly diagnostic of their planning strategies \cite{Callaway2017}. In this paradigm, participants plan the route of a spider through a maze with the goal to maximise their score (see Figure \ref{fig:mouselab}), which is the sum of the value of the nodes (grey circles) along the path they chose to traverse. Each node has a gain or loss that is initially concealed but may be seen by clicking on it. This explicit clicking motion indicates that the person is judging the quality of a hypothetical future state, which we regard as an elementary planning operation. The operation's cognitive cost is externalised by charging a price that varies depending on the experimental condition. Participants are therefore incentified to reveal information based on the cost and reward of planning. In this way, the paradigm externalises the mental model that people use to plan in terms of which and how many nodes have been clicked.

\begin{minipage}{0.45\textwidth}
\begin{figure}[H]
    \centering
    \includegraphics[width=0.6\linewidth]{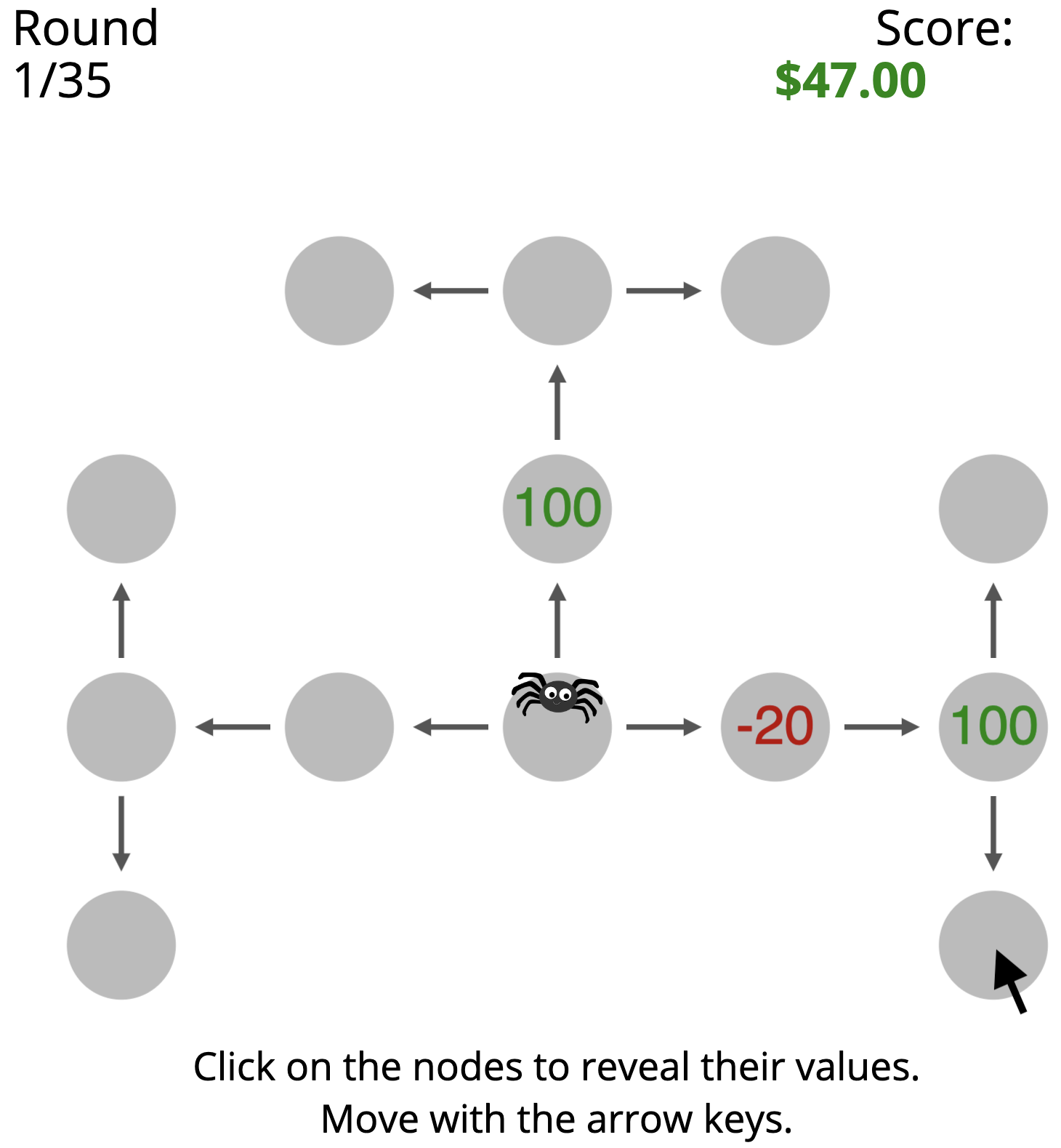}
    \caption{Screenshot example of one trial of the Mouselab paradigm for the high reward variance low click cost condition (HVLC) with three nodes revealed} 
    \label{fig:mouselab}
\end{figure}
\end{minipage}
\hfill
\begin{minipage}{0.45\textwidth}
\begin{table}[H]
    \centering
    \scriptsize
    \begin{tabular}{lll}
    \hline
                                                              & \begin{tabular}[c]{@{}l@{}}High reward \\ variance\end{tabular}                        & \begin{tabular}[c]{@{}l@{}}Low reward \\ variance\end{tabular}                        \\ \hline
    \begin{tabular}[c]{@{}l@{}}High click\\ cost\end{tabular} & \begin{tabular}[c]{@{}l@{}}High variance \\ High cost condition \\ (HVHC)\end{tabular} & \begin{tabular}[c]{@{}l@{}}Low variance \\ High cost condition \\ (LVHC)\end{tabular} \\ \hline
    \begin{tabular}[c]{@{}l@{}}Low click \\ cost\end{tabular} & \begin{tabular}[c]{@{}l@{}}High variance \\ Low cost condition \\ (HVLC)\end{tabular}  & \begin{tabular}[c]{@{}l@{}}Low variance \\ Low cost condition \\ (LVLC)\end{tabular}  \\ \hline
    \end{tabular}
    \caption{Experiment settings and corresponding condition abbreviations. Experimental conditions are the four possible combinations of two sets of possible range of rewards (high and low variance) and two levels of planning costs (high and low)}
    \label{table:conditions}
\end{table}
\end{minipage}

\subsection{Methods}
The experiment was conducted in accordance to study protocol 429/2021BO2 approved by the Independent Ethics Commission of the Medical Faculty of the University T\"ubingen. 

\paragraph{Materials} 
Using the Mouselab-MDP paradigm, we created an experiment that independently varies the benefit and the cost of planning across four conditions (see Table~\ref{table:conditions}). We manipulated the benefit of planning by defining two sets of available rewards. One set contains a large range of available rewards (i.e., the rewards are drawn uniformly from the set \{-1000, -100, -50, 50, 100\}), whereas in the other set, the difference between the available rewards is small (i.e., the rewards are drawn uniformly from the set \{-6, -4, -2, 2, 4, 6\}). We denote the conditions using the set with large differences as \textit{high-variance} conditions and the conditions using set with small differences \textit{low-variance} conditions. In addition, we manipulated the cost of planning by setting the click cost to either $-5$ (\textit{high-cost}) or $-1$ (\textit{low-cost}). 


\paragraph{Participants and procedure}
We recruited 208 participants on CloudResearch, that is 52 participants for each condition. The recruitment was limited to participants who had completed 100+ Human Intelligence Tasks (HIT), had HIT approval rate of at least 90\% \cite{cloudresearch}, and were located in the United States. The participants were randomly assigned to one of the four conditions and were given minimal instructions followed by a quiz to test their correct understanding. After passing the quiz, participants were asked to complete 35 trials of planning in the Mouselab-MDP paradigm. Participants could earn a performance-dependent bonus up to \$5 in addition to their base-pay of \$1.50
The scores are displayed on the screen and are updated after each click or move. Participants were informed that they would receive 0.2 cents for each point of their final score after completing all trials. The average bonus participants received was \$1.79 in the HVHC condition, \$2.08 in the HVLC condition, \$0.02 in the LVHC condition, and \$0.10 in LVLC condition. The HIT took on average 14 minutes. \footnote{The experiment can be tested here: \url{http://planning-amount.herokuapp.com}.} We excluded 15 participants (7\%) from the analysis because they did not engage in any planning (clicking) in any of the trials.

\subsection{Results} 
To examine whether people adapt the amount of planning to its cost and benefits, we analysed their number of clicks as the decision to click corresponds to a decision to plan more because additional planning is necessary to get any benefit out of the additional information. We hypothesised that in each condition participants would gradually learn to adapt the amount of planning they perform. Concretely, we predicted that in high-variance conditions the number of clicks would increase over time because the large range of possible rewards makes planning unusually beneficial and at the same time causing not planning to be very costly. By contrast, in low-variance conditions planning is less beneficial and therefore the number of clicks should decrease. Furthermore, we predicted that people would adapt to the high cost of planning by reducing their number of clicks more strongly when the benefit of planning is low and increasing it less strongly when the benefit of planning is high.

\begin{figure}[h!]
    \centering
    \begin{subfigure}[b]{0.48\textwidth}
        \caption{HVHC condition. Mann Kendall test suggests increasing trend ($S=237, p<.001$). Average number of clicks: $5.58$;}
        \includegraphics[width=0.8\textwidth,height=3.8cm]{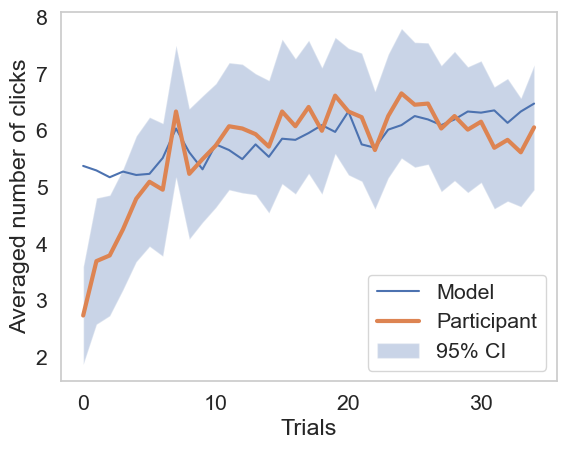}
        \label{fig:cond1}
    \end{subfigure}\hfill%
    ~ 
    \begin{subfigure}[b]{0.48\textwidth}
    \caption{HVLC condition. Mann Kendall test suggests increasing trend ($S=429, p<.0001$). Average number of clicks: $5.69$}
        \includegraphics[width=0.8\textwidth,height=3.8cm]{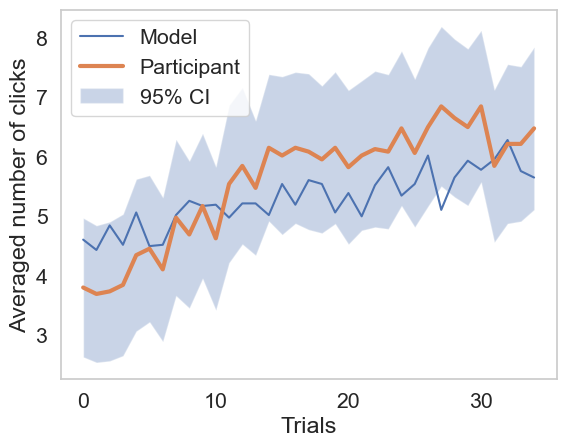}
        \label{fig:cond2}
    \end{subfigure}
    \centering
    \begin{subfigure}[b]{0.48\textwidth}
        \caption{LVHC condition. Mann Kendall test suggests decreasing trend ($S=-473, p<.0001$). Average number of clicks: $0.64$}
        \includegraphics[width=0.8\textwidth,height=3.8cm]{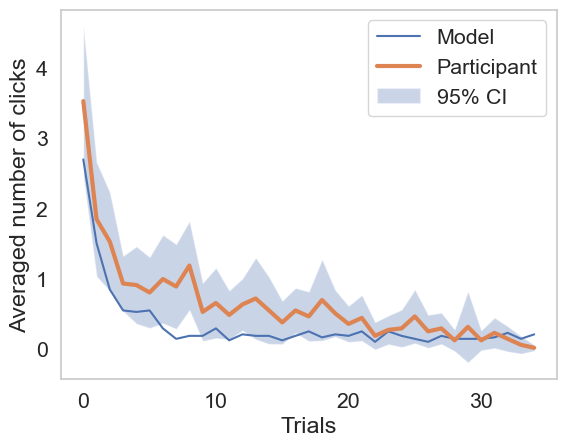}
        \label{fig:cond3}
    \end{subfigure}\hfill%
    \begin{subfigure}[b]{0.48\textwidth}
        \caption{LVLC condition. Mann Kendall test suggests decreasing trend ($S=-287, p<.0001$). Average number of clicks in the LVLC condition: $2.59$}
        \includegraphics[width=0.8\textwidth,height=3.8cm]{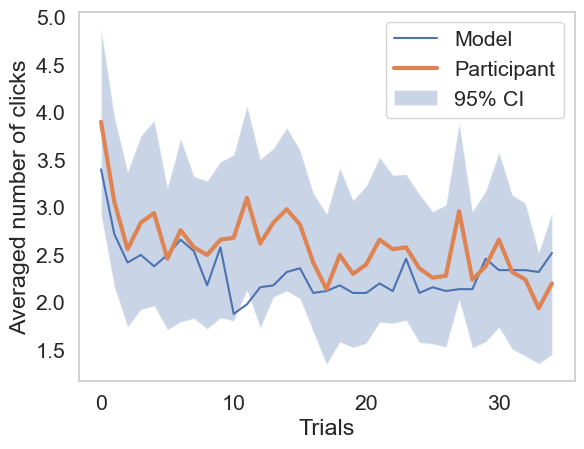}
        \label{fig:cond4}
    \end{subfigure}
    \caption{Averaged click development of all participants and of the fitted model (REINFORCE with metacognitive pseudo-rewards)} 
    \label{fig:averagedclickdevelopment}
\end{figure}

\begin{figure}[h!]
    \centering
    \includegraphics[width=0.3\textwidth]{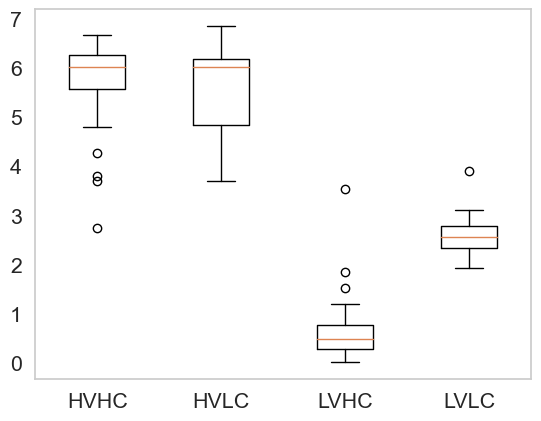}
    \caption{Box plot of participants' numbers of clicks for all four conditions.}
    \label{fig:boxplot}
\end{figure}

To test our hypotheses, we first visually examined the click development across the trials. Figure \ref{fig:averagedclickdevelopment} shows the averaged number of clicks across participants for all four conditions. The number of clicks increased significantly in the high-variance conditions, where planning is highly beneficial, and decreased significantly in the low-variance conditions, where planning is less beneficial. 
A significant Kruskal-Wallis ANOVA showed that the four groups differed in how much they planned ($S=14.66, p=.002$). Pair-wise Wilcoxon rank sum tests comparing the average numbers of clicks suggested significant differences between the low-variance conditions ($S=-6.79, p<.0001$), the high-cost conditions ($S=7.18, p<.0001$) and the low-cost conditions ($S=7.14, p<.0001$) but not for the high-variance conditions ($S=0.21, p=.83$). These results as well as the box plot (see Figure~\ref{fig:boxplot}) suggest that people do learn to rationally adapt their amount of planning to its cost and benefit. In addition, the benefit of planning in the high-variance conditions by far outweighs the cost of planning and therefore the participants learn to plan more regardless of the costs, whereas the cost of planning makes a difference in the low-variance conditions, where the benefit of planning is already marginal and the benefits could easily be outweighed by the costs.

\section{Modelling metacognitive learning}
Having shown that people do learn to adapt their amount of planning, we then model the underlying metacognitive learning mechanism by applying \textit{reinforcement learning} algorithms to the problem of deciding how much to plan (\textit{meta-decision-making}) \cite{Boureau2015,Griffiths2019}. We now briefly introduce these two frameworks and how they can be combined.

\subsection{Background}

\subsubsection{Reinforcement learning}
According to previous work, human learning is driven by incentives and punishments received through trial and error \cite{niv2009reinforcement}. Reinforcement learning algorithms are based on this learning mechanism as they learn to estimate how much reward may be expected from taking a particular action $a$ in a given state $s$. This estimate is updated according to the differences between received and predicted rewards $\delta$, the reward prediction errors: 
\begin{equation}
    Q(s,a) \leftarrow Q(s,a) - \alpha \cdot \delta 
\end{equation}
where $Q$ denotes the Q-value \cite{watkins1992q} and $\alpha$ is the learning rate. To strike a compromise between exploitation and exploration, the agent can pick its actions  \textit{probabilistically}, maximising the predicted action value, for example using the softmax rule \cite{Williams1992}:
\begin{equation}
    P(a|s,Q) \propto \exp(\nicefrac{1}{\tau} \cdot Q(s,a)) \label{eq:softmaxPolicy}
\end{equation}
where $\tau$ is the inverse temperature parameter.

\subsubsection{Meta-decision-making}
According to previous research, the brain is equipped with multiple decision systems that interact in a variety of ways \cite{Dolan2013,Daw2018}. The model-based system, in contrast to Pavlovian and model-free systems, allows for flexible reasoning about which action is preferable 
but demands a process for deciding which information should be considered in a given decision. Therefore, an important part of deciding how to decide is to efficiently balance decision quality and decision time given a huge amount of information. This is known as \emph{meta-decision-making} \cite{Boureau2015}. 
Recent work has formalised the problem of meta-decision-making as a meta-level MDP \cite{krueger2017enhancing,Griffiths2019}:
\begin{equation}
M_{meta}=\left( \mathcal{B}, \mathcal{C} \cup \lbrace \bot \rbrace, T_{meta}, r_{meta} \right),\label{eq:metaLevelMDP} \\
Q_{meta} (b, c) \leftarrow Q_{meta} (b, c) - \alpha \cdot \delta_{meta}
\end{equation} 
where belief states $b_t \in \mathcal{B}$ encode the model-based decision system's beliefs about the values of alternative courses of actions. The temporal evolution of those belief states ($b_1,b_2,\cdots$) is driven by the decision system's computations $c_1,c_2,\cdots$ according to the meta-level transition probabilities $T(b_t,c_t,c_{t+1})$. Finally, the meta-level reward function $r_{\text{meta}}(b_t,c_t)$ encodes the cost of performing the planning operation $c_t\in \mathcal{C}$ and the expected return of terminating planning ($c_t=\bot$) and acting based on the current belief state $b_t$. This meta-level MDP can for example by solved by applying meta Q-learning, that is $Q_{meta}(b, c)$. 

\subsubsection{Metacognitive reinforcement learning}
Planning strategies can be viewed as policies for resolving MDPs at the metalevel. Hence, the problem of identifying efficient planning methods can be formalised as solving a metalevel MDP for the best metalevel policy \cite{Griffiths2019}. Although it is often computationally hard to solve meta-decision-making problems optimally, the best solution can be approximated by reinforcement learning \cite{Russell1991,callaway2018learning}. As a result, we assume that the brain approximates optimal meta-decision-making through reinforcement learning mechanisms that attempt to approximate the optimal solution of the meta-level MDP defined in Equation \ref{eq:metaLevelMDP} by either learning to approximate the optimal policy directly \cite{HeJainLieder2021} or learning an approximation to its value function \cite{jain2019people}. Previous research has used this concept to describe how people learn to choose between different cognitive strategies \cite{Erev2005,rieskamp2006ssl,LiederGriffiths2017}, how many steps to plan ahead \cite{krueger2017enhancing}, when to exercise how much cognitive control \cite{LiederShenhav2018} and how people learn which information to consider \cite{HeJainLieder2021}. This approach, however, has yet to be applied to the study of how people develop and refine their cognitive strategies with respect to how much information to consider for planning depending on its cost and benefit.

\subsection{Models of metacognitive reinforcement learning}
To examine the underlying metacognitive learning mechanism, we build on previous work on metacognitive reinforcement learning 
According to \cite{jain2019people}, a value-based RL model called the \textit{Learned Value of Computation} (LVOC) model seems to be able to explain how people learn planning strategies reasonably well. On the other hand, work by \cite{HeJainLieder2021} suggests that people's adaptation of their planning strategy to different environment structures might follow a policy-gradient mechanism called REINFORCE that is additionally supported by internally generated metacognitive rewards for generating valuable information. Furthermore, work by \cite{jain2019people} suggests that human planning might be controlled by two sequential meta-control decisions about whether planning should be continued (Stage 1) and, if so, which planning operation should be performed next (Stage 2). 
To examine which mechanism, value-based or policy-gradient reinforcement learning, best explains how people learn how much to plan, we test existing models of those mechanisms on our new data. In addition, we extend both the value-based and policy-gradient models by adding internally generated metacognitive rewards for generating valuable information as well as hierarchical approaches. \footnote{The code accompanying this work can be found here \url{https://github.com/Reeche/planningamount}.} 

\subsubsection{Representations of the planning strategies}
Our models represent people's planning strategies as softmax policies operating on a weighted combination of 52 neuroscientifically informed features (see Appendix~\ref{appendix:featureslist}). One example of a group of features is pruning features \cite{Huys2012}, which are related to assigning a negative value to think about a path whose expected value is below a certain threshold. With this representation, a person's learning trajectory can be described as a time series of the weight vectors that represent the person's planning strategies in terms of those features.

\subsubsection{REINFORCE model}
According to the REINFORCE model \cite{jain2019people}, which is based on the REINFORCE algorithm \cite{Williams1992}, people change their planning strategy directly by updating their softmax policy (see Equation~\ref{eq:softmaxPolicy}).  
After each decision, the weight vector $\theta$ is updated in the direction of the gradient of the difference between its the returns of its choices and the cost of the performed planning operations:
\begin{equation}
    \theta \leftarrow \theta + \alpha \cdot \sum_{t=1}^{O}\gamma^{t-1} \cdot r_{meta}(b_t,c_t) \cdot \nabla_\theta \ln \pi_\theta(c_t | b_t), \label{eq:reinforceUpdate}
\end{equation}
where $b$ represents the belief state, $c$ the click under consideration, $\mathcal{C}_{b}$ the set of clicks available in belief state $b$, $\alpha$ the learning rate, $\gamma$ the discount factor, $\mathbf{f}$ the above-mentioned features and $O$ is the number of planning operations executed by the model on that trial. The learning rate $\alpha$ was optimised using ADAM \cite{kingma2014adam}. Next to the initial weight vector $\theta$, the REINFORCE model has three free parameters: $\alpha$, $\gamma$ and inverse temperature $\tau$ that are fit separately for each participant.

\subsubsection{LVOC model}
According to the LVOC model, people identify and adjust their strategy continuously by learning to anticipate the values of alternative planning operations \cite{krueger2017enhancing}. This is achieved by approximating the meta-level Q-function by a linear combination of the features mentioned above:
\begin{equation}
	Q_{\text{meta}}(b_k,c_k) \approx \sum_{j=1}^{52} w_j \cdot f_j(b_k,c_k),\label{eq:FunctionApproximation}
\end{equation}
The weights of those features are learned by Bayesian linear regression of the bootstrap estimate $\hat{Q}(b_k,c_k)$ of the meta-level value function onto the features $\mathbf{f}$:
\begin{equation}
	\hat{Q}(b_k,c_k)=r_{\text{meta}}(b_k,c_k)+ \langle \mu_t , \mathbf{f}(b',c') \rangle \label{eq:bootstrap}
\end{equation}
The sum of the immediate meta-level reward and the anticipated value of the future belief state $b'$ under the present meta-level policy is the bootstrap estimate in Equation \ref{eq:bootstrap}.
The predicted value of $b'$ is the scalar product of the the posterior mean $\mu_t$ of the weights $\mathbf{w}$ given the observations from all preceding planning operations and the features $\mathbf{f}(b',c')$ of $b'$ and the cognitive operation $c'$ that the current policy picks given state. A generalised variant of Thompson sampling selects the next planning operation $c'$ based on the posterior on the feature weights $\mathbf{w}$. That means, to make the $k$\textsuperscript{th} meta-decision, $n$ weight vectors $\tilde{w}_1,\cdots,\tilde{w}_n$ are sampled from the posterior distribution of the weights given the series of meta-level states, selected planning operations, and resulting value estimates encountered so far: 
\begin{equation}
	\tilde{w}^{(1)}_k, \cdots, \tilde{w}^{(n)}_k \sim P(\mathbf{w} | \mathcal{E}_k),
\end{equation}
where the set $\mathcal{E}_k=\lbrace e_1, \cdots, e_k \rbrace$ contains the meta-decision-maker's experience from the first $k$ meta-decisions. To be precise, each meta-level experience $e_j\in\mathcal{E}_k$ is a tuple $\left(b_j,h_j,\hat{Q}(b_j,c_j; \mu_j)\right)$ containing a meta-level state, the computation selected in it, and the bootstrap estimates of its Q-value. The arithmetic mean of the sampled weight vectors $\tilde{w}^{(1)},\cdots,\tilde{w}^{(n)}$ is then used to predict the Q-values of each potential planning operation $c \in \mathcal{C}$ according to Equation~\ref{eq:FunctionApproximation}. 
The model then either exploits what it has learned so far by choosing the planning operation with the highest predicted Q-value or explores a random planning operation with probability $p$.
The LVOC model therefore has the following free parameters: $p$, the mean vector $\mu_{prior}$ and variance $\sigma^2_{\text{prior}}$ of its prior distribution $\mathcal{N}(\mathbf{w};\mu_{prior}, \sigma^2\cdot\text{I} )$ on the weights $\mathbf{w}$, and the number of samples $n$. 

\subsubsection{Metacognitive features}
We augmented the REINFORCE and LVOC models with two components: metacognitive rewards for generating valuable information and a two-stage hierarchical meta-decision-making process.

\paragraph{Metacognitive rewards for generating valuable information}
Because of the central role of reward prediction errors in reinforcement learning \cite{Schultz1997,Glimcher2011} and the scarcity of external rewards in metacognitive reinforcement learning \cite{Hay2016}, we postulate that the brain might accelerate this learning process by generating additional metacognitive pseudo-rewards that convey the value of the information generated by the just performed planning operation. Concretely, the value of the pseudo-reward for performing computation $c_t$ in belief state $b_t$ and transitioning to belief state $b_{t+1}$ is:
\begin{equation}
    \text{PR}(b_t,c,b_{t+1})=\mathds{E}[R_{\pi_{b_{t+1}}}|b_{t+1}]-\mathds{E}[R_{\pi_{b_t}}|b_{t+1}], \label{eq:PR}
\end{equation}
which is the difference between the expected value of the best path in belief state $b_{t+1}$ according to the policy $\pi_{b_{t+1}}$ and the expected value of the best path in belief state $b_{t+1}$ according to the behaviour policy $\pi_{b_t}$ (e.g., moving along the path up-up-right through the environment shown in Figure~\ref{fig:mouselab}) which is defined as
\begin{equation}
    \pi_{b}(s) = \argmax_a E_b [R | s,a] 
\end{equation}
where $R$ is the sum of the external rewards (e.g., the sum of rewards collected by moving through the maze) and the expected value is taking with respect to the probability distribution encoded by $b$.

\paragraph{Hierarchical meta-control}
Previous research suggests that foraging decisions are made by two distinct decision systems: the ventromedial prefrontal cortex and the dorsal anterior cingulate cortex \cite{Rushworth2012}. Since determining how and how much to plan is similar to foraging for information, it might also rely on two separate systems \cite{jain2019people}. We therefore, developed an extension to the LVOC and REINFORCE models that first decides whether to continue planning (Stage 1) and then selects the next planning operation according to either the LVOC model or the REINFORCE model (Stage 2). Stage 1 utilises an adaptive satisficing stopping rule \cite{Callaway2018}, which adjusts the satisficing threshold based on the number of clicks made through a free parameter $\eta$. The smaller $\eta$ the more likely it is to terminate planning. The hierarchical approach therefore, introduced one additional free parameter $\eta$ to the base models LVOC and REINFORCE. 


\subsection{Model fitting methods}
The resulting 8 different models were assessed on how well they can capture how people learn how much to plan. For that, we fitted each model's free parameters to the participant's data and applied each model to the series of problems the participant had to solve. The parameters were fit by maximising a multivariate-Normal pseudo-likelihood function defined in terms of the probability that the model would generate the participant's trial wise number of clicks $\mathbf{c}$ as a function of its parameters: 
For a given participant $i$, the pseudo-likelihood function under model $m$ is given by:
\begin{equation}
    \mathcal{L}\left( (\theta_{i,m},\sigma_{i,m}) | \mathbf{c_i}\right) = \phi(\mathbf{c_i}; \mathbf{\hat{c}_{i,m}}(\theta), \sigma_{i,m} I) \label{eq:pseudoLikelihood}
\end{equation} 
where $\theta_{i,m}$ is the parameter vector used to fit the data from participant $i$ with model $m$, $\mathbf{c_{i}}$ is the vector of number of clicks that the $i$\textsuperscript{th} participant performed on trials 1 through 35, $\sigma$ is the standard deviation of the errors between the observed number of clicks and the model's predictions $\mathbf{\hat{c}_{i,m}}(\theta_{i,m})$, and $\phi(\mathbf{x}; \mu, \Sigma)$ is the density function of the multivariate normal distribution. We estimate the parameters $\theta_{i,m}$ and $\sigma_{i,m}$ by maximising the pseudo-likelihood function in Equation~\ref{eq:pseudoLikelihood} using Bayesian Optimisation \cite{bergstra2013making}. All 8 combinations of the LVOC and REINFORCE models, that is with or without pseudo-reward and hierarchical as well as non-hierarchical variants are then fit to the participant data using 400 iterations. In each iteration, the model's prediction is estimated by averaging the model's scores across 30 simulations.


\subsection{Model selection}
After the model-fitting, we performed model selection using the Bayesian information criterion (BIC) \cite{schwarz1978estimating} and Bayesian model selection. Concretely, we use random effects Bayesian model selection \cite{rigoux2014bayesian,stephan2009bayesian} at the group level to estimate the expected proportion of people that are best described by a given model ($r$) and the so-called \textit{exceedance} probability $\phi$ that this proportion is significantly higher than the corresponding proportion for any other model. In addition, we perform family-level Bayesian model selection to draw the equivalent inferences for sets of models that share a common feature \cite{penny2010comparing}. 
In general, REINFORCE models appear to be superior to LVOC models with 137 out of 193 participants best explained by REINFORCE variants according to the BIC. Bayesian model selection on a model-family level supported this conclusion by suggesting that approximately $77.35\%$ of the population can be best described by REINFORCE models and this proportion is higher than the proportion of people that are best described by any of the other models ($\phi>99\%$). In addition, our data suggested that models with and without pseudo-rewards are about equally good at explaining metacognitive learning ($r_{\text{PR}}=49.63\%$, $\phi_{\text{PR}}=46.91\%$). Furthermore, most participants' learning behaviour did not support the hierarchical meta-control mechanism ($r_{\text{HR}}=41.40\%$, $\phi_{\text{HR}}=1.63\%$). For more details, see Tables~\ref{table:countofbestfit}, \ref{table:bmsmodel}, \ref{table:bmspr}, and \ref{table:bmshr} in Appendix~\ref{appendix:modelselection}. 

According to Bayesian model selection at the level of individual models, we can be 95\% confident that the REINFORCE model with pseudo-rewards explains the learning behaviour of a greater proportion of people than any of the alternative models ($r=25.10\%$, $\phi=88.07\%$; see Table~\ref{table:bmsoverall}). Counting the number of participants best fitted by a given model according to the BIC corroborated this conclusion. The REINFORCE model with pseudo-reward also had the lowest average BIC value (i.e., $142.91$). The fits for the participants of the REINFORCE model with pseudo-rewards are shown in Figure~\ref{fig:averagedclickdevelopment}. Details as well as the figures of the averaged performance of all the other models can be found in Table~\ref{table:bic}, Figures~\ref{fig:model1855details_cond4} and Figures~\ref{fig:allotherplots} in the appendix. Some shortcomings of the model are for example the flat increase in the number of clicks at the beginning of the HVHC condition, the lower click amount level in HVLC condition and a steeper decrease in the number of clicks in the LVHC condition. 

\begin{table}[h!]
\centering
\scriptsize
\caption{Results of Bayesian model selection comparing all 8 combinations of the models and metacognitive features, namely the vanilla LVOC model (LVOC), the LVOC model that uses pseudo-rewards (LVOC-PR), the LVOC model with hierarchical meta-control (HR-LVOC), the hierarchical LVOC model that uses pseudo-rewards (HR-LVOC-PR), and the corresponding variants of the REINFORCE model (RF). 
}
\begin{tabular}{lllllllll}
\hline
 &  LVOC & LVOC-PR & HR-LVOC & HR-LVOC-PR & RF & \textbf{RF-PR} & HR-RF & HR-RF-PR \\ \hline
Proportion ($r$)  & 7.44\%    & 7.74\%    & 6.24\%   &  2.31\%  & 18.13\%    & \textbf{25.10\%}   & 14.87\%  & 18.17\%  \\
Exceedance prob. ($\phi$) & 0\%    & 0\%    & 0\%   & 0\%  & 5.62\%   & \textbf{88.07\%}   & 0.61\%   & 5.70\%  
\\ \hline
\end{tabular}
\label{table:bmsoverall}
\end{table}

\subsection{Model-based analysis}
Participants differed in their ability to adapt their amount of planning. To investigate those differences, we divided them into three groups: participants in the high-variance conditions whose number of clicks significantly increased according to Mann Kendall tests of trend or participants in the low-variance conditions whose number of clicks significantly decreased are classified as \textit{highly adaptive}, participants with significantly decreasing numbers of clicks in the high-variance conditions and participants with significantly increasing numbers of clicks in the low-variance conditions are classified as \textit{maladaptive}, and the other participants are classified as \textit{moderately adaptive}. To gain insights into the learning processes of each group, we performed a model-based analysis using the REINFORCE model with pseudo-reward as it best explained a larger proportion of participants' data than the alternative models. 

We hypothesised that maladaptive participants would have different learning rates than the other two groups and tested this hypothesis using Wilcoxon rank-sum tests on the fitted learning rates. For the condition with low reward variance and low click cost, the maladaptive learners had an average learning rate of $0.025$, while the highly adaptive learners had an average learning rate of $0.006$ and the moderately adaptive learners had an average learning rate of $0.010$.  The tests implied that the distribution of the fitted learning rate differs significantly between maladaptive and highly adaptive participants ($M=2.02, p=.04$) as well as between maladaptive and moderately adaptive participants ($M=2.34, p=.02$). 
The high learning rates of maladaptive learners might have manifested in the high volatility of their average number of clicks shown in the appendix  Figure~\ref{fig:model1855details_cond4}. This suggests that their high learning rates might have led to overshooting the target and oscillating back and forth between suboptimal extremes. 
Bonferroni corrected Wilcoxon rank-sum tests did not show any other statistically significant findings for any of the other model parameters or conditions. 


\section{Conclusion and further work}
Meta-control over decision-making is an important aspect of human metacognition. To investigate how such metacognitive decisions are shaped by learning, we measured how people adapt how much they plan to its costs and benefits and then modelled the underlying learning mechanisms in terms of metacognitive reinforcement learning. Using a process-tracing method, we found that the amount of planning, defined in the number of clicks, increased significantly in conditions, where planning is highly beneficial and decreased significantly in conditions, where planning is less beneficial. The cost of clicking also had a significant effect in the less beneficial conditions as participants planned more when the cost is low and planned less when the cost of planning is high. After having confirmed that people learned to adapt their amount of planning depending on its cost and benefits, we proceeded to test a set of metacognitive reinforcement learning models enhanced by metacognitive features.  Model selection using BIC and Bayesian model selection suggests that participants might rely on a policy-gradient mechanism that generates its own metacognitive pseudo-rewards. In addition, high learning rates was discovered to be one potential source of maladaptiveness.

In summary, our findings suggest that the metacognitive decision of whether to do more planning is partly made through metacognitive reinforcement learning from past experience. This finding provides additional support to the emerging view that metacognitive reinforcement learning plays an important role in people's metacognitive ability to adapt their decision strategies to the requirements of their environment \cite{jain2019people,LiederGriffiths2017,LiederShenhav2018,krueger2017enhancing}. This suggests that developing metalevel reinforcement learning algorithms \cite{Hay2016,callaway2018learning,lieder2014algorithm} is a promising avenue to recreating this ability in machines. 
This work mainly focused on the amount of planning externalised by the number of clicks. Future work should investigate which model best predicts which planning operations people perform (cf. \cite{HeJainLieder2021}).



\bibliographystyle{abbrv}
\bibliography{neurips_2021}

\appendix

\section{Appendix}

\subsection{Features description} \label{appendix:featureslist}

\subsubsection{Mental effort avoidance}
\textbf{Feature 1: ``Termination Constant"}:
The value of this feature is 1 for all clicks and 0 for the termination operation in all belief states.

\subsubsection{Model-based metareasoning features}
These features capture uncertainty about the values of the unobserved nodes. Uncertainty is defined as the standard deviation of the values of the distribution. The following features capture uncertainty:\\
\\
\textbf{Feature 2: ``Uncertainty"}: The value of this feature for a click in a given belief state is the uncertainty in the value of the considered node.
\\
\\
\textbf{Feature 3: ``Max Uncertainty"}: The value of this feature for a click in a given belief state is the is the maximum uncertainty in return for the current trial from all the paths that the considered node lies on.
\\
\\
\textbf{Feature 4: ``Successor Uncertainty"}: The value of this feature for a click in a given belief state is the total uncertainty in the values of all the successors of the considered node on the current trial.
\\
\\
\textbf{Feature 5: ``Trial level standard deviation"}: The value of this feature for a click is the uncertainty in the value of the considered node as estimated across all trials attempted so far by the agent.
\\
\\
\textbf{Feature 6: ``Current trial level standard deviation"}: The value of this feature for a click in a given belief state is the uncertainty in the value of nodes at the same depth as the considered node as estimated for the current trial.
\\
\\
\textbf{Feature 7: ``Does the node lie on the second most promising path?"}: The value of this feature for a click in a given belief state is 1 if the considered node lies on the path which has the second highest expected return for the current trial, and 0 otherwise.

\subsubsection{Pavlovian Features} These features are based on greedy maximization. Pavlovian behavior is captured by the following features:
\\
\\
\textbf{Feature 8: ``Best expected value"}: The value of this feature for a click in a given belief state is the best expected return for a path in the current trial among all the paths that the considered node lies on.
\\
\\
\textbf{Feature 9: ``Best largest value"}: The value of this feature for a click in a given belief state is the maximum value observed among all the paths that the considered node lies on.
\\
\\
\textbf{Feature 10: ``Does the node lie on the most promising path?"}: The value of this feature for a click in a given belief state is 1 if the considered node lies on the path with the highest expected return for the current trial, and 0 otherwise.
\\
\\
\textbf{Feature 11: ``Value of the max expected return"}: The value of this feature for all clicks in a given belief state is the maximum expected return from all paths in the current trial.
\\
\\
\textbf{Feature 12: ``Does a successor node have a maximum value?"}: The value of this feature for a click in a given belief state is 1 if any of the considered node's observed successors in the current trial has a value which is the maximum possible value for the reward distribution, and 0 otherwise.
\\
\\
\textbf{Feature 13: ``Maximum value of a successor"}: The value of this feature for a click in a given belief state is the maximum value that has been observed among all the successors of the considered node in the current trial.
\\
\\
\textbf{Feature 14: ``Maximum value of an immediate successor"}: The value of this feature for a click in a given belief state is the maximum value that has been observed among all the immediate successors of the considered node in the current trial.
\\
\\
\textbf{Feature 15: ``Value of the parent node"}: The value of this feature for a click in a given belief state is the value of the considered node's parent if the parent node has been observed, and 0 otherwise.

\paragraph{Pruning features}
These features are designed to capture the idea of pruning branches \cite{Huys2012}. The value for these features for all clicks is -1 if the maximum expected loss that can be incurred in the current belief state is worse than the pruning threshold and 0 otherwise. We consider features with different pruning thresholds (features 16-19). In addition, we consider the following features:\\
\\
\textbf{Feature 20: ``Soft Pruning"}: The value of this feature for a clicks is the maximum expected loss that can be incurred in a given belief state from all paths that the considered node lies on.
\\
\\
\textbf{Feature 21: ``Is the previous observed node a successor and has negative value"}: The value of this feature for a click in a given belief state is 1 if the last observed node in the current trial is a child of the considered node and has a negative value, and 0 otherwise.

\subsubsection{Satisficing and stopping features}

\paragraph{Satisficing features} These features determine when the planning satisfices \cite{simon1956rational}. The value for these features is -1 for termination if the maximum expected return for the current trial is greater than the satisficing threshold. We consider features with different satisficing thresholds (features 22-28). In addition, we consider the following 2 features:
\\
\\
\textbf{Feature 29: ``Soft Satisficing"}: The value of this feature for all clicks in a given belief state is the maximum return that can be expected on the current trial from all paths that the considered node lies on.

\paragraph{Stopping Criteria}
These features have same value for all the clicks and a different value for termination.\\
\\
\textbf{Feature 30: ``Are all max paths observed?"}: The value of this feature is -1 for all clicks and 0 for termination action in all belief states if all the paths path leading to a final outcome, which has the maximum value among the observed final outcomes, has been observed in the current trial and 0 otherwise.
\\
\\
\textbf{Feature 31: ``Is a max path observed?"}: The value of this feature is -1 for all clicks in all belief states if any path leading to the node, which has the maximum value possible for the reward distribution, has been observed in the current trial and 0 otherwise.
\\
\\
\textbf{Feature 32: ``Is a positive node observed?"}: The value of this feature is -1 for all clicks in all belief states if a node with a positive value has been observed in the current trial and 0 otherwise.
\\
\\
\textbf{Feature 33: ``Is the previous observed node maximal?"}: The value of this feature is -1 for all clicks if the last observed node in the current trial has the maximum value possible for the reward distribution and 0 otherwise.
\\
\\
\textbf{Feature 34: ``Is a complete path observed?"}: The value of this feature is -1 for all nodes in all belief states if at least one path has been completely observed from immediate outcome to final outcome, and 0 otherwise.
\\
\\
\textbf{Feature 35: ``All final outcomes observed?"}: The value of this feature is -1 for all clicks in all belief states if all final outcomes have been observed in the current trial and 0 otherwise.
\\
\\
\textbf{Feature 36: ``Are all immediate outcomes observed?"}: The value of this feature is -1 for all clicks in all belief states if all immediate outcomes have been observed in the current trial and 0 otherwise.
\\
\\
\textbf{Feature 37: ``Are final outcomes of positive immediate outcomes observed?"}: The value of this feature is -1 for all clicks in all belief states if all the final outcomes that can be reached from the positive observed immediate outcomes have been observed, and 0 otherwise.

\subsubsection{Model-free values and heuristics features}
\paragraph{Relational features}
The values of these features for a considered node are dependent on its neighboring nodes.\\
\\
\textbf{Feature 38: ``Ancestor count"}: The value of this feature for a click in a given belief state is the number of ancestors of the considered node that have been observed in the current trial.
\\
\\
\textbf{Feature 39: ``Depth Count"}: The value of this feature for a click in a given belief state is the number of times that any node at the same depth as the considered node has been observed in the current trial.
\\
\\
\textbf{Feature 40: ``Is the node a final outcome and has a positive ancestor?"}: The value of this feature for a click in a given belief state is 1 if the considered node is a final outcome and it has an observed ancestor with a positive value in the current trial and 0 otherwise.
\\
\\
\textbf{Feature 41: ``Immediate successor count"}: The value of this feature for a click in a given belief state is the number of children of the considered node that have been observed in the current trial.
\\
\\
\textbf{Feature 42: ``Is parent observed?"}: The value of this feature for a click in a given belief state is 1 if the parent node of the considered node has been observed, and 0 otherwise.
\\
\\
\textbf{Feature 43: ``Successor Count"}: The value of this feature for a click in a given belief state is the number of observed successors of the considered node for the current trial.
\\
\\
\textbf{Feature 44: ``Squared Successor Count"}: The value of this feature for a click in a given belief state is the square of the number of observed successors of the considered node for the current trial.
\\
\\
\textbf{Feature 45: ``Siblings Count"}: The value of this feature for a click in a given belief state is the number of siblings of the considered node that have been observed in the current trial.
\\
\\
\textbf{Feature 46: ``Minimum number of observed nodes on branch"}: The value of this feature for a click in a given belief state is the minimum number of nodes observed on all the branches containing the considered node.
\\
\\
\textbf{Feature 47: "Is the previous observed node a successor?"}: The value of this feature for a click in a given belief state is 1 if the last observed node in the current trial is one of the successors of the considered node, and 0 otherwise.

\paragraph{Structural features}
The values of these features are dependent no the task structure.\\
\\
\textbf{Feature 48: ``Depth"}: The value of this feature for a click in a given belief state is the distance of the considered node from the starting position.
\\
\\
\textbf{Feature 49: ``Is the node an immediate outcome?"}: The value of this feature for a click in a given belief state is 1 if the considered node in an immediate outcome and 0 otherwise.
\\
\\
\textbf{Feature 50: ``Is the node a final outcome?"}:The value of this feature for a click in a given belief state is 1 if the considered node is a final outcome and 0 otherwise.
\\
\\
\textbf{Feature 51: "Observed height"}: The value of this feature for a click in a given belief state is the length of the maximum observed path to a final outcome starting from the considered node.

\subsection{Model performance}

\subsubsection{Model selection} \label{appendix:modelselection}

\paragraph{Bayesian information criterion (BIC)}
First we counted the number of models best fitted to individual participants according to the lowest BIC. Then we grouped the participants into highly adaptive, moderately adaptive and maladaptive participants and calculated the average BIC according to the groups. 
\begin{table}[h!]
\centering
\scriptsize
\caption{Count of fitted individual participants’model with lowest BIC. LVOC corresponds to the vanilla LVOC. LVOC-PR means LVOC model that uses pseudo-reward. HR-LVOC denotes the hierarchical LVOC variant. HR-LVOC-PR is the hierarchical LVOC model that uses pseudo-reward. The same applies to the REINFORCE model which we abbreviate as RF.}
\begin{tabular}{lllllllll}
\hline
\textbf{}               & LVOC & PR-LVOC & HR-LVOC & HR-PR-LVOC & RF         & PR-RF       & HR-RF & HR-PR-RF    \\ \hline
HVHC condition          & 2    & 3       & 5       & 1          & 6          & \textbf{17} & 10    & 6           \\
HVLC condition          & 5    & 7       & 3       & 0          & 10         & 6           & 4     & \textbf{11} \\
LVHC condition & 6    & 3       & 6       & 3          & \textbf{9} & 8           & 7     & 5           \\
LVLC condition          & 5    & 5       & 1       & 1          & 8          & \textbf{13} & 6     & 11          \\ \hline
Sum                     & 18   & 18      & 15       & 5          & 33         & \textbf{44} & 27    & 33          \\ \hline
\end{tabular}
\label{table:countofbestfit}
\end{table}

\begin{table}[h!]
\centering
\scriptsize
\caption{Averaged BIC for each model grouped by participants and averaged within conditions. Best performance is marked in bold}
\begin{tabular}{lllllllll}
\hline
                        & LVOC   & RF-LVOC & HR-LVOC & HR-PR-LVOC & RF              & PR-RF           & HR-RF  & HR-PR-RF        \\ \hline
HVHC condition          & 186.44 & 185.77  & 186.44  & 186.79     & 182.83          & \textbf{174.77} & 177.32 & 180.91          \\
Highly adaptive (n=16)  & 193.19 & 193.12  & 188.94  & 189.33     & 192.60          & 188.04          & 180.65 & \textbf{180.23} \\
Maladaptive (n=11       & 172.29 & 178.01  & 182.34  & 183.10     & 169.56          & \textbf{162.56} & 172.40 & 181.06          \\
Mod. adaptive (n=23)    & 188.50 & 184.37  & 186.66  & 186.79     & 182.38          & \textbf{171.39} & 177.35 & 181.31          \\ \hline
HVLC condition          & 169.49 & 173.59  & 180.06  & 176.15     & \textbf{159.70} & 163.24          & 163.69 & 164.27          \\
Highly adaptive (n=26)  & 186.95 & 188.64  & 191.22  & 189.07     & \textbf{181.71} & 187.48          & 182.28 & 183.29          \\
Maladaptive (n=6)       & 150.83 & 146.23  & 163.60  & 150.12     & 133.86          & 134.23          & 131.15 & \textbf{130.56} \\
Mod. adaptive (n=14)    & 145.05 & 157.37  & 166.38  & 163.30     & \textbf{129.91} & 130.66          & 143.12 & 143.38          \\ \hline
LVHC condition & 93.33  & 100.26  & 111.13  & 110.50     & 95.44           & \textbf{92.98}  & 101.96 & 100.83          \\
Highly adaptive (n=27)  & 96.05  & 99.68   & 112.54  & 110.17     & \textbf{93.58}  & 94.29           & 103.60 & 98.25           \\
Maladaptive (n=0)       & \multicolumn{8}{c}{No maladaptive participants}                                                        \\
Mod. adaptive (n=20)    & \textbf{89.67}  & 101.04  & 109.23  & 110.94     & 97.96           & 91.21  & 99.73  & 104.31          \\ \hline
LVLC condition          & 147.71 & 144.73  & 145.93  & 147.36     & 139.20          & 140.65          & 140.17 & \textbf{136.79} \\
Highly adaptive (n=9)   & 147.26 & 138.83  & 144.53  & 145.43     & \textbf{129.53} & 133.50          & 139.84 & 137.76          \\
Maladaptive (n=8)       & 156.56 & 150.78  & 151.67  & 151.08     & 146.55          & 145.54          & 152.26 & \textbf{141.42} \\
Mod. adaptive (n=33)    & 145.68 & 144.87  & 144.92  & 146.99     & 140.06          & 141.42          & 137.33 & \textbf{135.40}
\end{tabular}
\label{table:bic}
\end{table}

\newpage 

\paragraph{Bayesian model selection}
Family-level Bayesian model selection was performed to compare different family of models: LVOC vs. REINFORCE models; models that uses pseudo-reward vs. models that do not use pseudo-reward and hierarchical models vs. non-hierarchical models. 


\begin{table}[h!]
\centering
\caption{LVOC models vs. REINFORCE models}
\begin{tabular}{lll}
\hline
             & Proportion ($r$)                                               & Exceedance probability ($\phi$)                                          \\ \hline  
HVHC condition & \begin{tabular}[c]{@{}l@{}}LVOC: 17.41\%, \\ RF: 82.59\%\end{tabular} & \begin{tabular}[c]{@{}l@{}}LVOC: \textless{}1\%, \\ RF: \textgreater{}99\%\end{tabular} \\
HVLC condition & \begin{tabular}[c]{@{}l@{}}LVOC: 24.74\%, \\ RF: 75.26\%\end{tabular} & \begin{tabular}[c]{@{}l@{}}LVOC: 00.04\%, \\ RF: 99.96\%\end{tabular}                   \\
LVHC condition & \begin{tabular}[c]{@{}l@{}}LVOC: 37.74\%, \\ RF: 62.26\%\end{tabular} & \begin{tabular}[c]{@{}l@{}}LVOC: 10.42\%, \\ RF: 89.58\%\end{tabular}                   \\
LVLC condition & \begin{tabular}[c]{@{}l@{}}LVOC: 17.54\%, \\ RF: 82.46\%\end{tabular} & \begin{tabular}[c]{@{}l@{}}LVOC: \textless{}1\%, \\ RF: \textgreater{}99\%\end{tabular} \\ \hline
Overall        & \begin{tabular}[c]{@{}l@{}}LVOC: 22.65\%, \\ RF: 77.35\%\end{tabular} & \begin{tabular}[c]{@{}l@{}}LVOC: \textless{}1\%, \\ RF: \textgreater{}99\%\end{tabular}\\ \hline
\end{tabular}
\label{table:bmsmodel}
\end{table}


\begin{table}[h!]
\centering
\caption{Pseudo-reward models vs. No pseudo-reward models}
\begin{tabular}{lll}
\hline
               & Proportion ($r$)                                                       & Exceedance probability ($\phi$)                                        \\ \hline
HVHC condition & \begin{tabular}[c]{@{}l@{}}PR: 63.34\%, \\ No-PR: 36.66\%\end{tabular} & \begin{tabular}[c]{@{}l@{}}PR: 95.03\%, \\ No-PR: 4.97\%\end{tabular}  \\
HVLC condition & \begin{tabular}[c]{@{}l@{}}PR: 38.06\%, \\ No-PR: 61.94\%\end{tabular} & \begin{tabular}[c]{@{}l@{}}PR: 11.30\%, \\ No-PR: 88.70\%\end{tabular} \\
LVHC condition & \begin{tabular}[c]{@{}l@{}}PR: 50.08\%, \\ No-PR: 49.92\%\end{tabular} & \begin{tabular}[c]{@{}l@{}}PR: 50.00\%, \\ No-PR: 50.00\%\end{tabular} \\
LVLC condition & \begin{tabular}[c]{@{}l@{}}PR: 49.28\%, \\ No-PR: 50.72\%\end{tabular} & \begin{tabular}[c]{@{}l@{}}PR: 46.55\%, \\ No-PR: 53.45\%\end{tabular} \\ \hline
Overall        & \begin{tabular}[c]{@{}l@{}}PR: 49.63\%, \\ No-PR: 50.37\%\end{tabular} & \begin{tabular}[c]{@{}l@{}}PR: 46.91\%, \\ No-PR: 53.09\%\end{tabular} \\ \hline
\end{tabular}
\label{table:bmspr}
\end{table}


\begin{table}[h!]
\centering
\caption{Hierarchical models vs. Non hierarchical models}
\begin{tabular}{lll}
\hline
               & Proportion ($r$)                                                       & Exceedance probability ($\phi$)                                        \\ \hline
HVHC condition & \begin{tabular}[c]{@{}l@{}}HR: 45.31\%, \\ No-HR: 54.69\%\end{tabular} & \begin{tabular}[c]{@{}l@{}}HR: 26.33\%, \\ No-HR: 73.67\%\end{tabular} \\
HVLC condition & \begin{tabular}[c]{@{}l@{}}HR: 38.09\%, \\ No-HR: 61.91\%\end{tabular} & \begin{tabular}[c]{@{}l@{}}HR: 6.94\%, \\ No-HR: 93.06\%\end{tabular}  \\
LVHC condition & \begin{tabular}[c]{@{}l@{}}HR: 40.74\%, \\ No-HR: 59.26\%\end{tabular} & \begin{tabular}[c]{@{}l@{}}HR: 14.84\%, \\ No-HR: 85.16\%\end{tabular} \\
LVLC condition & \begin{tabular}[c]{@{}l@{}}HR: 39.48\%, \\ No-HR: 60.52\%\end{tabular} & \begin{tabular}[c]{@{}l@{}}HR: 9.23\%, \\ No-HR: 90.77\%\end{tabular}  \\ \hline
Overall        & \begin{tabular}[c]{@{}l@{}}HR: 41.40\%, \\ No-HR: 58.60\%\end{tabular} & \begin{tabular}[c]{@{}l@{}}HR: 1.63\%, \\ No-HR: 98.37\%\end{tabular}  \\ \hline
\end{tabular}
\label{table:bmshr}
\end{table}

\newpage 

\subsubsection{Plots} \label{fig:allotherplots}
The following figures show the the performance of the remaining 7 different models. 

\begin{figure}[h!]
    \centering
    \begin{subfigure}[b]{0.48\textwidth}
        \includegraphics[width=0.8\textwidth]{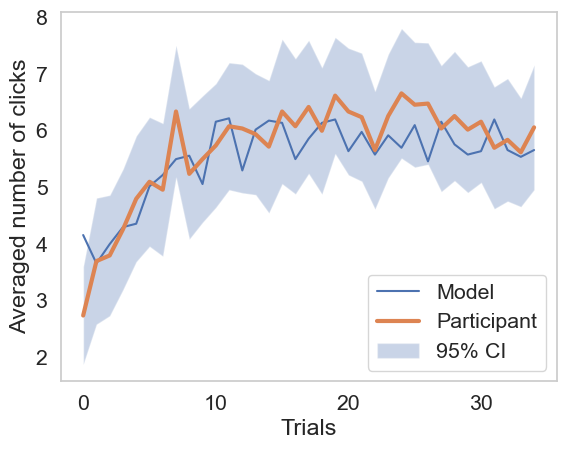}
        \caption{HVHC condition: high reward variance, high click cost}
    \end{subfigure}\hfill%
    \begin{subfigure}[b]{0.48\textwidth}
        \includegraphics[width=0.8\textwidth]{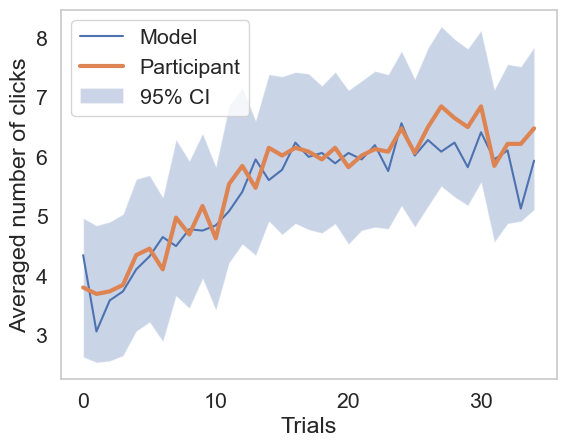}
        \caption{HVLC condition: high reward variance, low click cost}
    \end{subfigure}
    \centering
    \begin{subfigure}[b]{0.48\textwidth}
        \includegraphics[width=0.8\textwidth]{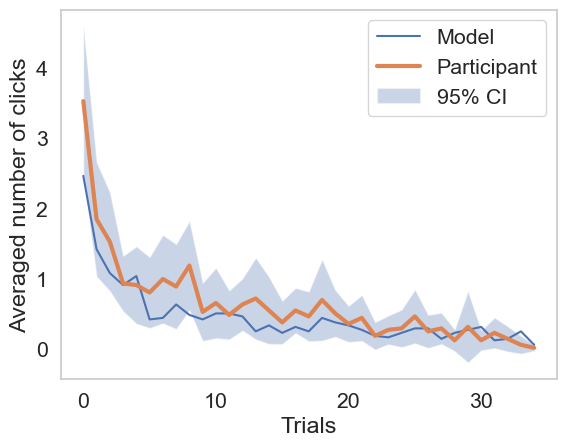}
        \caption{LVHC condition: low reward variance, high click cost}
    \end{subfigure}\hfill%
    \begin{subfigure}[b]{0.48\textwidth}
        \includegraphics[width=0.8\textwidth]{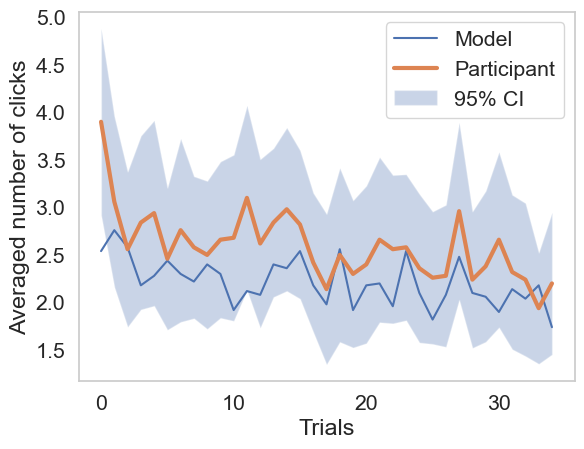}
        \caption{LVLC condition: low reward variance, low click cost}
    \end{subfigure}
    \caption{Averaged click development of vanilla LVOC }
\end{figure}

\begin{figure}[h!]
    \centering
    \begin{subfigure}[b]{0.48\textwidth}
        \includegraphics[width=0.8\textwidth]{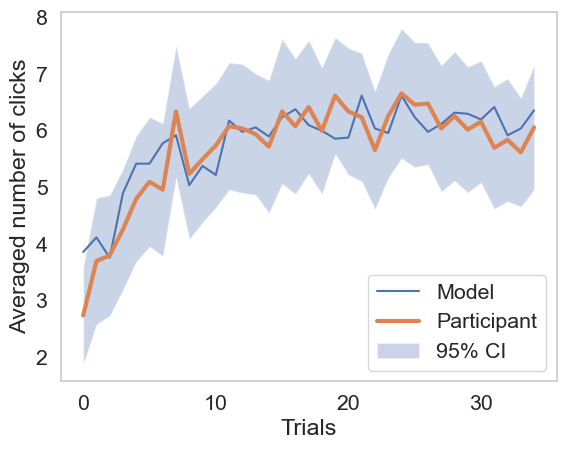}
        \caption{HVHC condition: high reward variance, high click cost}
    \end{subfigure}\hfill%
    \begin{subfigure}[b]{0.48\textwidth}
        \includegraphics[width=0.8\textwidth]{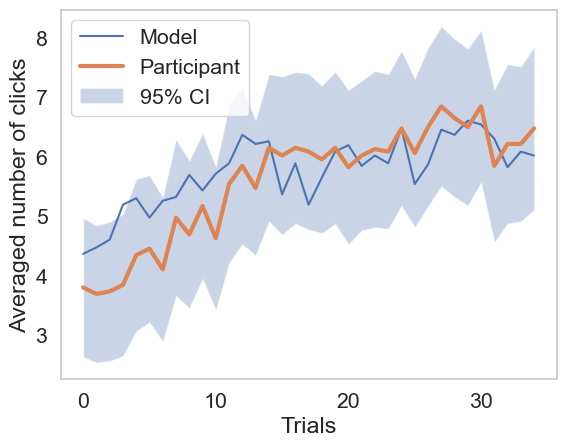}
        \caption{HVLC condition: high reward variance, low click cost}
    \end{subfigure}
    \centering
    \begin{subfigure}[b]{0.48\textwidth}
        \includegraphics[width=0.8\textwidth]{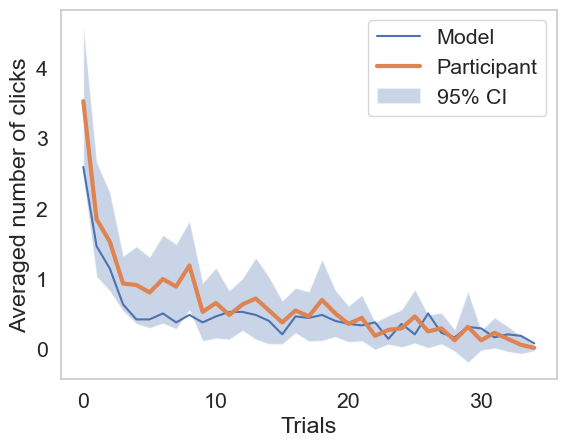}
        \caption{LVHC condition: low reward variance, high click cost}
    \end{subfigure}\hfill%
    \begin{subfigure}[b]{0.48\textwidth}
        \includegraphics[width=0.8\textwidth]{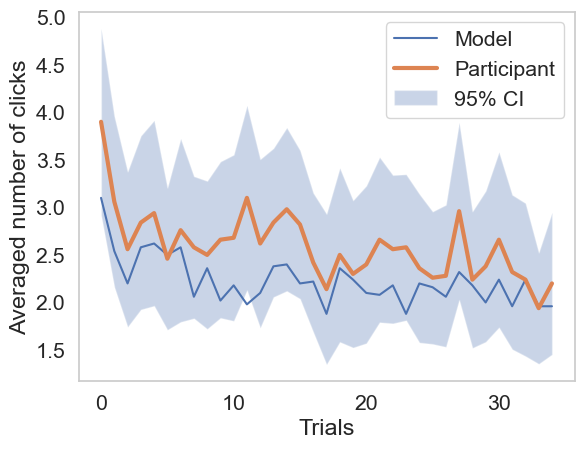}
        \caption{LVLC condition: low reward variance, low click cost}
    \end{subfigure}
    \caption{Averaged click development of LVOC with pseudo-reward}
\end{figure}

\begin{figure}[h!]
    \centering
    \begin{subfigure}[b]{0.48\textwidth}
        \includegraphics[width=0.8\textwidth]{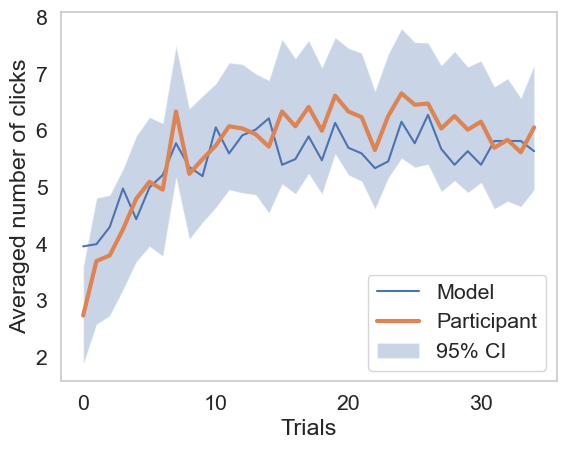}
        \caption{HVHC condition: high reward variance, high click cost}
    \end{subfigure}\hfill%
    ~ 
    \begin{subfigure}[b]{0.48\textwidth}
        \includegraphics[width=0.8\textwidth]{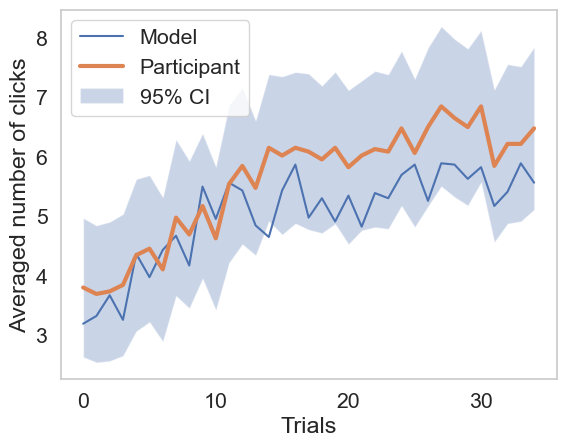}
        \caption{HVLC condition: high reward variance, low click cost}
    \end{subfigure}
    \centering
    \begin{subfigure}[b]{0.48\textwidth}
        \includegraphics[width=0.8\textwidth]{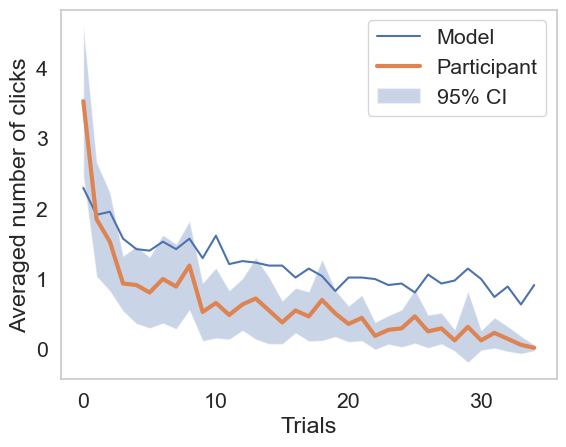}
        \caption{LVHC condition: low reward variance, high click cost}
    \end{subfigure}\hfill%
    \begin{subfigure}[b]{0.48\textwidth}
        \includegraphics[width=0.8\textwidth]{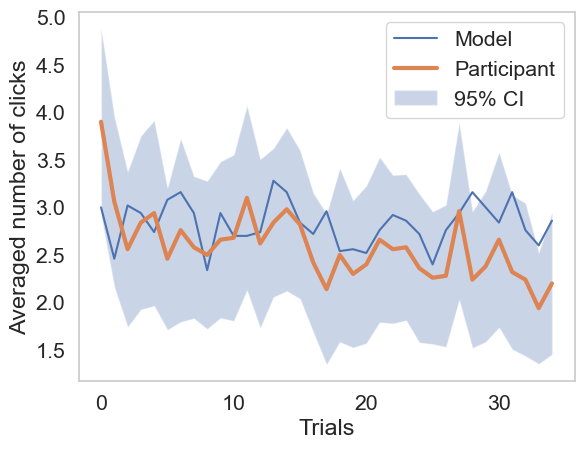}
        \caption{LVLC condition: low reward variance, low click cost}
    \end{subfigure}
    \caption{Averaged click development of hierarchical LVOC}
\end{figure}

\begin{figure}[h!]
    \centering
    \begin{subfigure}[b]{0.48\textwidth}
        \includegraphics[width=0.8\textwidth]{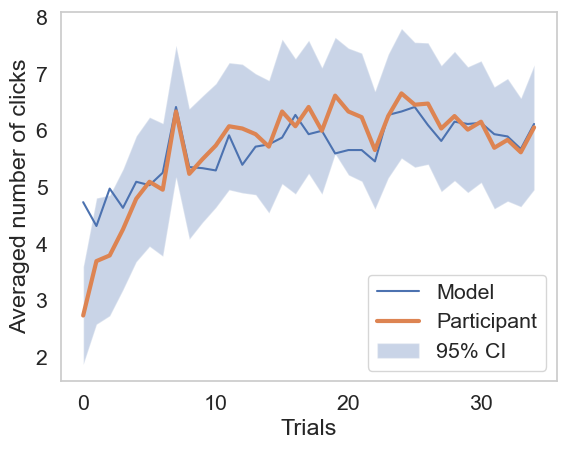}
        \caption{HVHC condition: high reward variance, high click cost}
    \end{subfigure}\hfill%
    ~ 
    \begin{subfigure}[b]{0.48\textwidth}
        \includegraphics[width=0.8\textwidth]{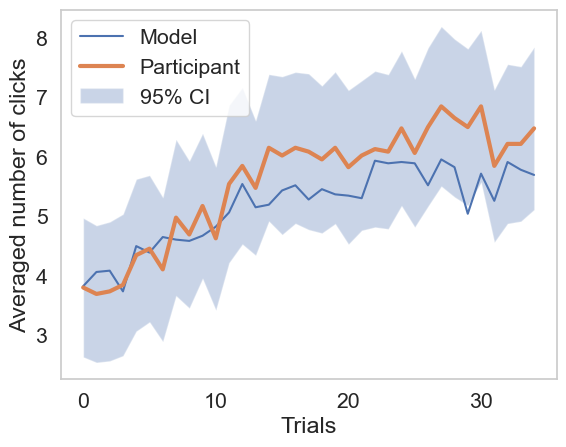}
        \caption{HVLC condition: high reward variance, low click cost}
    \end{subfigure}
    \centering
    \begin{subfigure}[b]{0.48\textwidth}
        \includegraphics[width=0.8\textwidth]{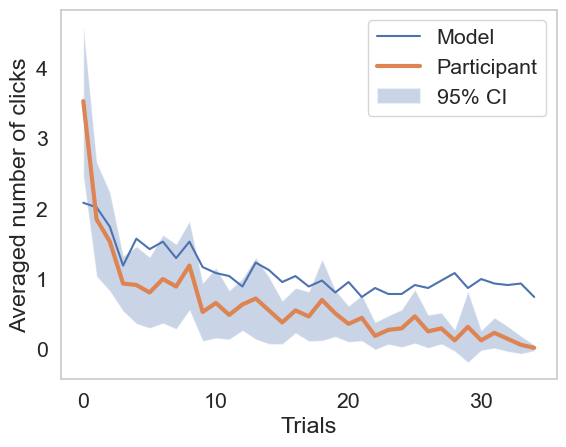}
        \caption{LVHC condition: low reward variance, high click cost}
    \end{subfigure}\hfill%
    \begin{subfigure}[b]{0.48\textwidth}
        \includegraphics[width=0.8\textwidth]{plots/models/high_variance_low_cost_number_of_clicks_likelihood_31_.png}
        \caption{LVLC condition: low reward variance, low click cost}
    \end{subfigure}
    \caption{Averaged click development of hierarchical LVOC with pseudo-reward}
\end{figure}

\begin{figure}[h!]
    \centering
    \begin{subfigure}[b]{0.48\textwidth}
        \includegraphics[width=0.8\textwidth]{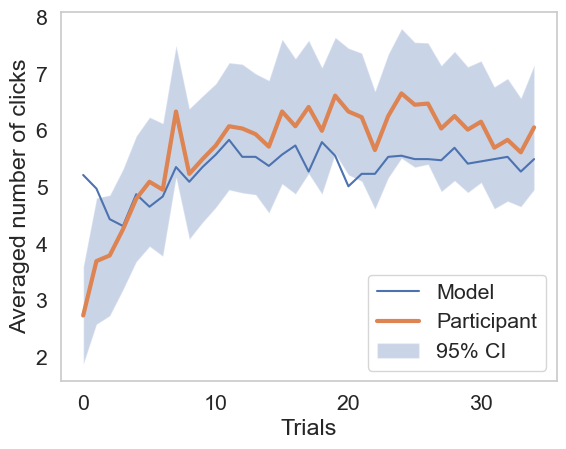}
        \caption{HVHC condition: high reward variance, high click cost}
    \end{subfigure}\hfill%
    ~ 
    \begin{subfigure}[b]{0.48\textwidth}
        \includegraphics[width=0.8\textwidth]{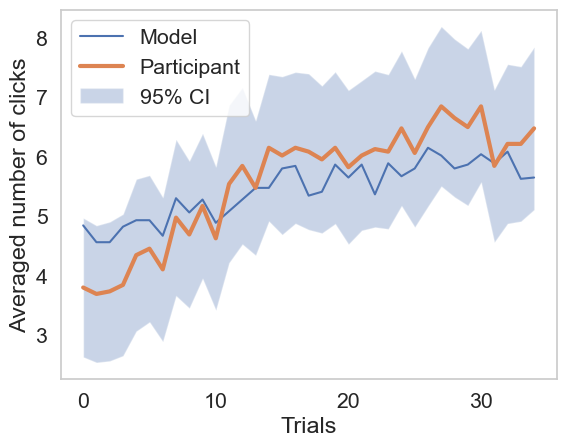}
        \caption{HVLC condition: high reward variance, low click cost}
    \end{subfigure}
    \centering
    \begin{subfigure}[b]{0.48\textwidth}
        \includegraphics[width=0.8\textwidth]{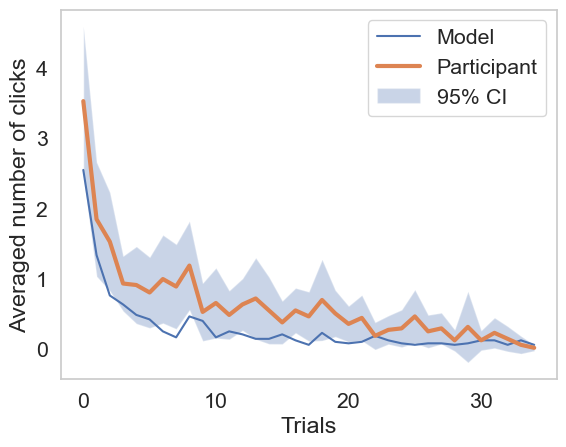}
        \caption{LVHC condition: low reward variance, high click cost}
    \end{subfigure}\hfill%
    \begin{subfigure}[b]{0.48\textwidth}
        \includegraphics[width=0.8\textwidth]{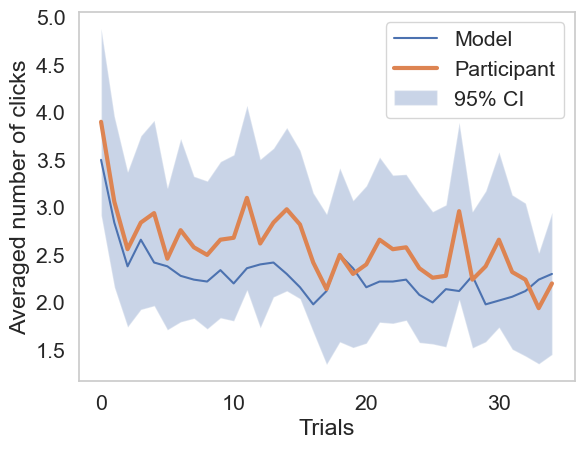}
        \caption{LVLC condition: low reward variance, low click cost}
    \end{subfigure}
    \caption{Averaged click development of vanilla REINFORCE }
\end{figure}

\begin{figure}[h!]
    \centering
    \begin{subfigure}[b]{0.48\textwidth}
        \includegraphics[width=0.8\textwidth]{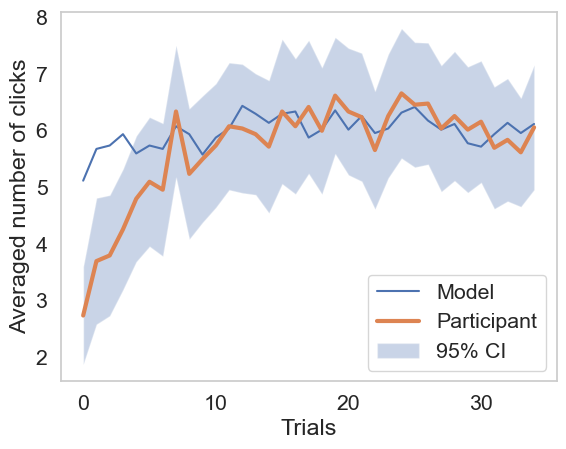}
        \caption{HVHC condition: high reward variance, high click cost}
    \end{subfigure}\hfill%
    ~ 
    \begin{subfigure}[b]{0.48\textwidth}
        \includegraphics[width=0.8\textwidth]{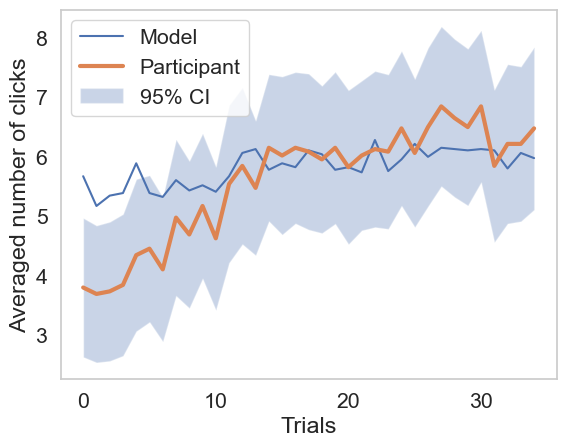}
        \caption{HVLC condition: high reward variance, low click cost}
    \end{subfigure}
    \centering
    \begin{subfigure}[b]{0.48\textwidth}
        \includegraphics[width=0.8\textwidth]{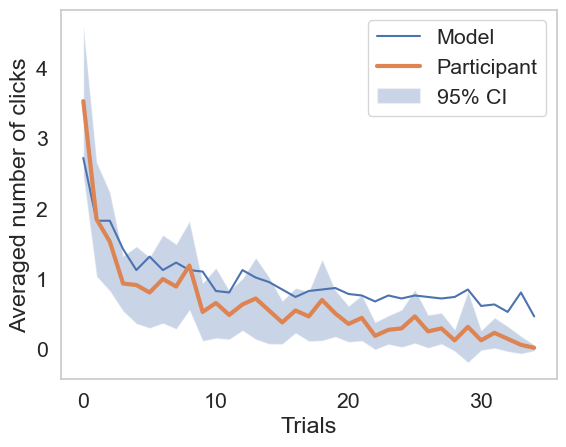}
        \caption{LVHC condition: low reward variance, high click cost}
    \end{subfigure}\hfill%
    \begin{subfigure}[b]{0.48\textwidth}
        \includegraphics[width=0.8\textwidth]{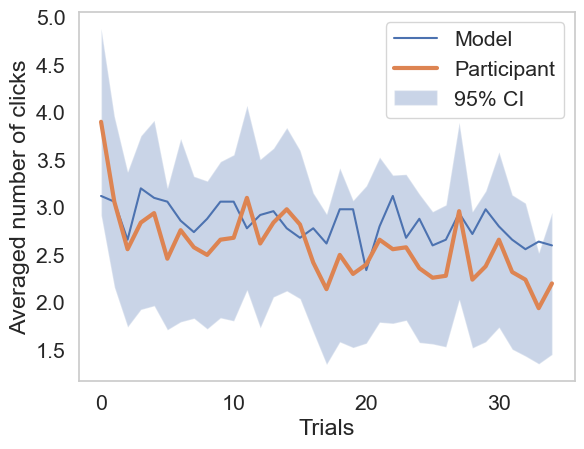}
        \caption{LVLC condition: low reward variance, low click cost}
    \end{subfigure}
    \caption{Averaged click development of hierarchical REINFORCE}
\end{figure}

\begin{figure}[h!]
    \centering
    \begin{subfigure}[b]{0.48\textwidth}
        \includegraphics[width=0.8\textwidth]{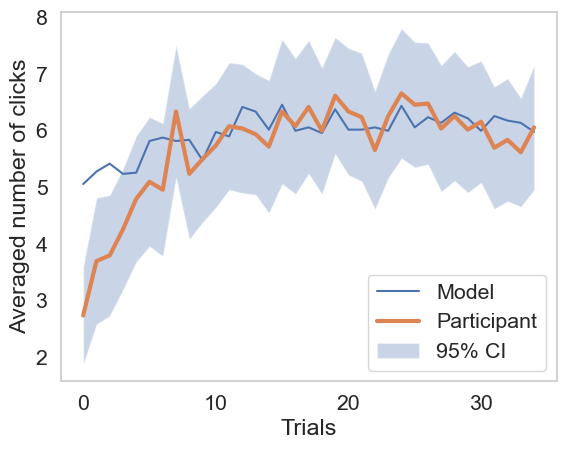}
        \caption{HVHC condition: high reward variance, high click cost}
    \end{subfigure}\hfill%
    ~ 
    \begin{subfigure}[b]{0.48\textwidth}
        \includegraphics[width=0.8\textwidth]{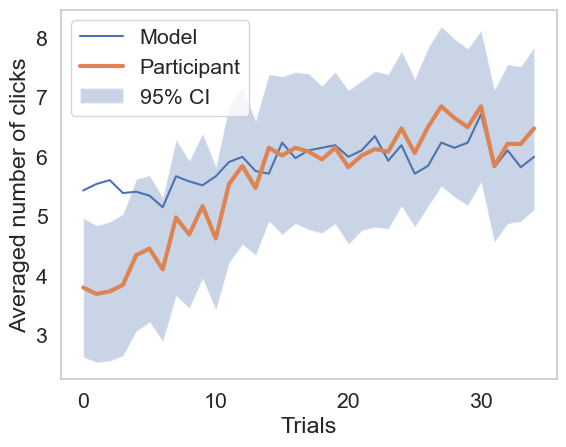}
        \caption{HVLC condition: high reward variance, low click cost}
    \end{subfigure}
    \centering
    \begin{subfigure}[b]{0.48\textwidth}
        \includegraphics[width=0.8\textwidth]{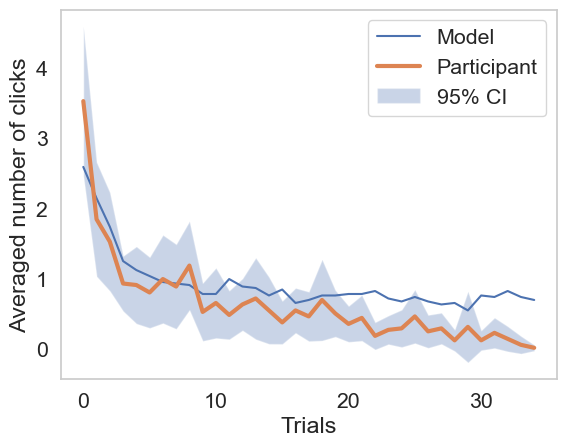}
        \caption{LVHC condition: low reward variance, high click cost}
    \end{subfigure}\hfill%
    \begin{subfigure}[b]{0.48\textwidth}
        \includegraphics[width=0.8\textwidth]{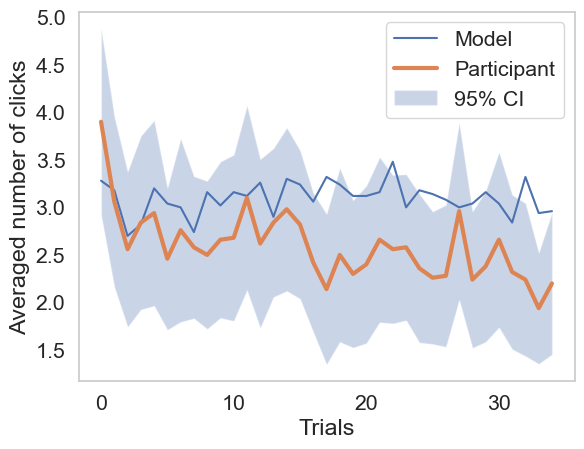}
        \caption{LVLC condition: low reward variance, low click cost}
    \end{subfigure}
    \caption{Averaged click development of hierarchical REINFORCE with metacognitive pseudo-rewards}
\end{figure}

\clearpage
\newpage

\subsection{Detailed model performance for REINFORCE with metacognitive pseudo-rewards}

\begin{figure}[h!]
    \centering
    \begin{subfigure}[b]{0.3\textwidth}
        \includegraphics[width=\textwidth]{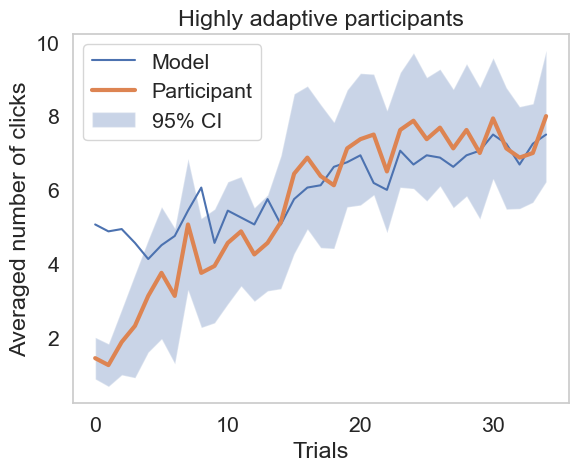}
    \end{subfigure}
    \begin{subfigure}[b]{0.3\textwidth}
        \includegraphics[width=\textwidth]{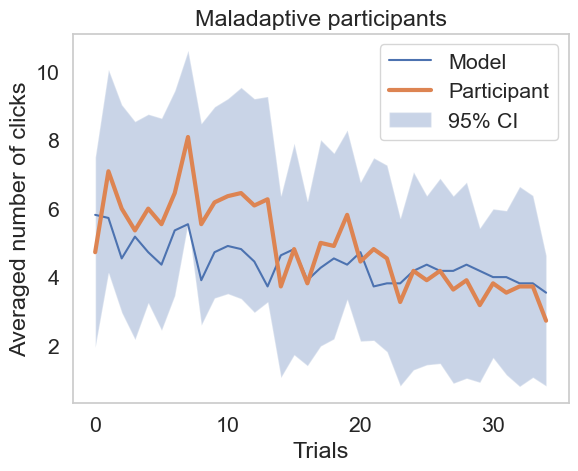}
    \end{subfigure}
    \begin{subfigure}[b]{0.3\textwidth}
        \includegraphics[width=\textwidth]{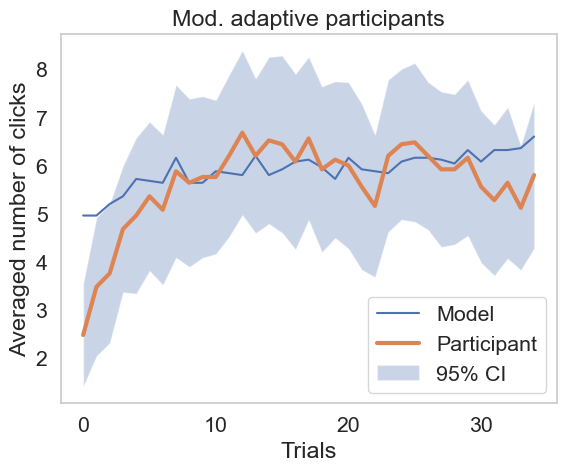}
    \end{subfigure}%
    
    \begin{subfigure}[b]{0.3\textwidth}
        \includegraphics[width=\textwidth]{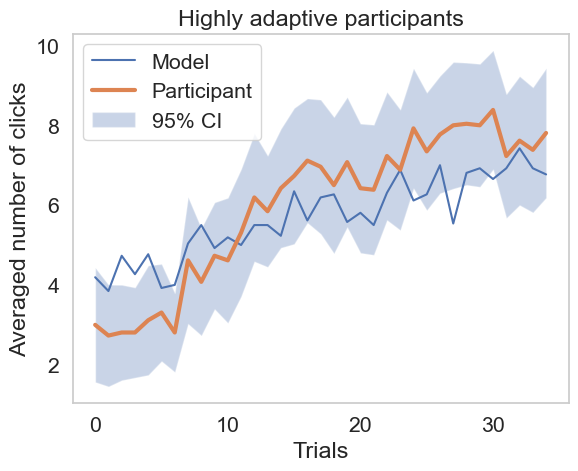}
    \end{subfigure}
    \begin{subfigure}[b]{0.3\textwidth}
        \includegraphics[width=\textwidth]{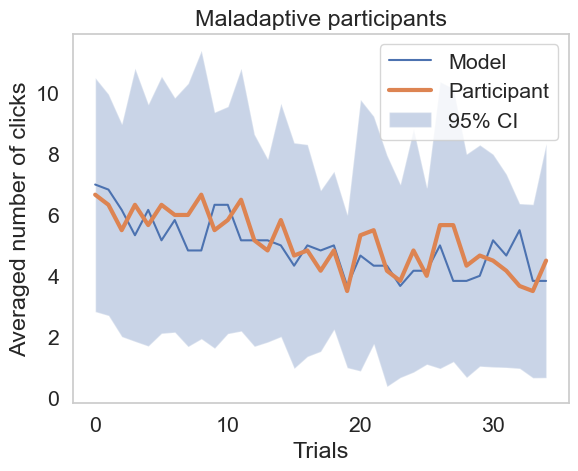}
    \end{subfigure}
    \begin{subfigure}[b]{0.3\textwidth}
        \includegraphics[width=\textwidth]{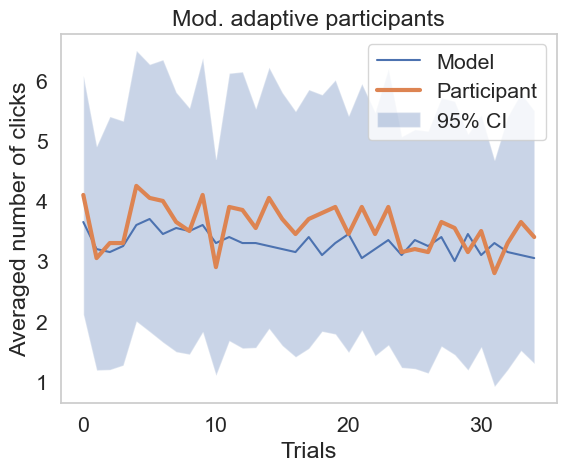}
    \end{subfigure}%
    
     \begin{subfigure}[b]{0.3\textwidth}
        \includegraphics[width=\textwidth]{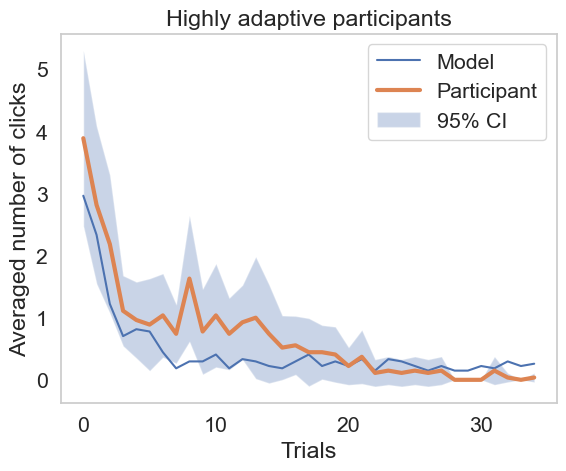}
    \end{subfigure}
    \begin{subfigure}[b]{0.3\textwidth}
        \includegraphics[width=\textwidth]{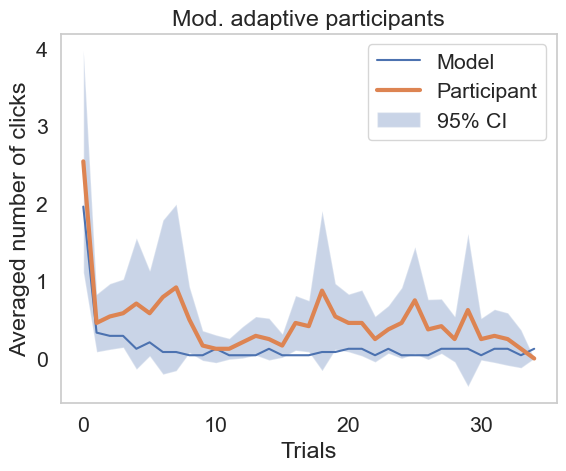}
    \end{subfigure}%
    
        \begin{subfigure}[b]{0.3\textwidth}
        \includegraphics[width=\textwidth]{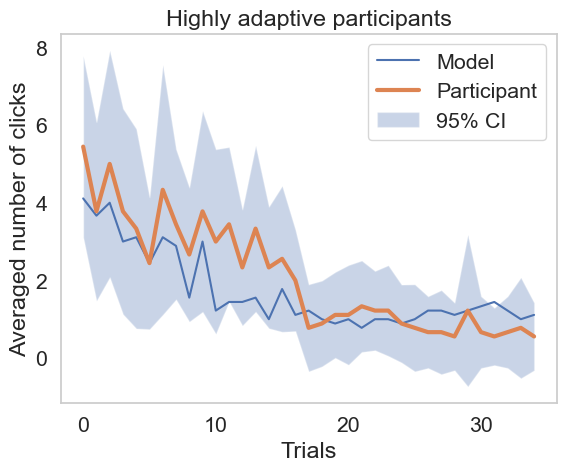}
    \end{subfigure}
    \begin{subfigure}[b]{0.3\textwidth}
        \includegraphics[width=\textwidth]{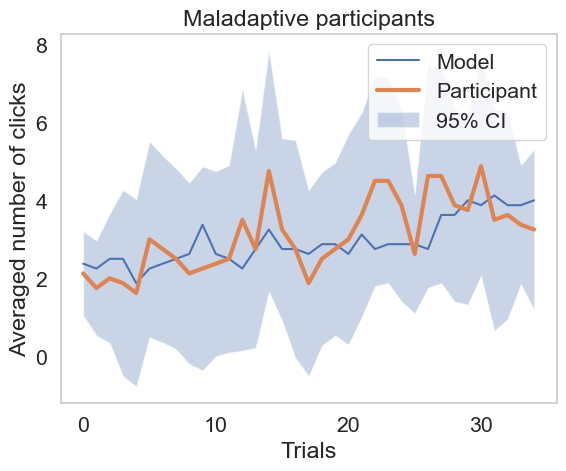}
    \end{subfigure}
    \begin{subfigure}[b]{0.3\textwidth}
        \includegraphics[width=\textwidth]{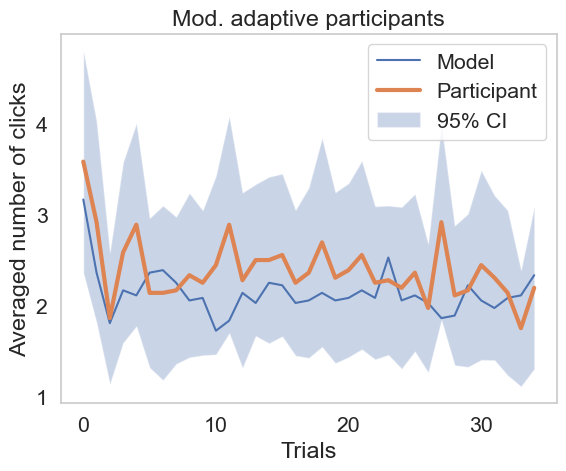}
    \end{subfigure}
    
    \caption{Averaged click development of the participants and of the fitted model (REINFORCE with metacognitive pseudo-rewards) for the HVHC, HVLC, LVHC and LVLC conditions} 
    \label{fig:model1855details_cond4}
\end{figure}

\end{document}